\documentclass[journal]{IEEEtran}
\ifCLASSINFOpdf
\else
\fi

\usepackage{amsmath,amsfonts,amssymb}
\usepackage{algpseudocode}

\usepackage[ruled,linesnumbered]{algorithm2e}  
\usepackage{array}
\usepackage{textcomp}
\usepackage{stfloats}
\usepackage{url}
\usepackage{verbatim}
\usepackage{graphicx}
\usepackage{caption}
\usepackage{subfigure}
\usepackage{cite}
\usepackage{multirow}
\usepackage{threeparttable}
\usepackage{hyperref}
\usepackage{multirow}
\usepackage{longtable}
\usepackage{xcolor}
\usepackage[table,xcdraw]{xcolor}
\usepackage{tablefootnote}
\usepackage{enumitem}
\usepackage[T1]{fontenc} 
\def\BibTeX{{\rm B\kern-.05em{\sc i\kern-.025em b}\kern-.08em
    T\kern-.1667em\lower.7ex\hbox{E}\kern-.125emX}}

\begin{document}
%
\title{EvoSpeak: Large Language Models for Interpretable Genetic Programming-Evolved Heuristics}
%
%
%

\author{Meng~Xu,
        Jiao~Liu,
        and~Yew~Soon~Ong,~\IEEEmembership{Fellow,~IEEE}
\thanks{Meng Xu is with the Singapore Institute of Manufacturing Technology, Agency for Science, Technology and Research, Singapore (e-mail: xu\_meng@simtech.a-star.edu.sg). Jiao Liu is with the College of Computing \& Data Science, Nanyang Technological University, Singapore (e-mail: jiao.liu@ntu.edu.sg). Yew Soon Ong is with the College of Computing and Data Science, Nanyang Technological University, and the Centre for Frontier AI Research, Institute of High Performance Computing, Agency for Science, Technology and Research, Singapore (e-mail: asysong@ntu.edu.sg).}
}

\markboth{Journal of \LaTeX\ Class Files,~Vol.~14, No.~8, August~2015}%
{Shell \MakeLowercase{\textit{et al.}}: Bare Demo of IEEEtran.cls for IEEE Journals}
%



\maketitle


\begin{abstract}
Genetic programming (GP) has demonstrated strong effectiveness in evolving tree-structured heuristics for complex optimization problems. Yet, in dynamic and large-scale scenarios, the most effective heuristics are often highly complex, hindering interpretability, slowing convergence, and limiting transferability across tasks. To address these challenges, we present EvoSpeak, a novel framework that integrates GP with large language models (LLMs) to enhance the efficiency, transparency, and adaptability of heuristic evolution. EvoSpeak learns from high-quality GP heuristics, extracts knowledge, and leverages this knowledge to (i) generate warm-start populations that accelerate convergence, (ii) translate opaque GP trees into concise natural-language explanations that foster interpretability and trust, and (iii) enable knowledge transfer and preference-aware heuristic generation across related tasks. We verify the effectiveness of EvoSpeak through extensive experiments on dynamic flexible job shop scheduling (DFJSS), under both single- and multi-objective formulations. The results demonstrate that EvoSpeak produces more effective heuristics, improves evolutionary efficiency, and delivers human-readable reports that enhance usability. By coupling the symbolic reasoning power of GP with the interpretative and generative strengths of LLMs, EvoSpeak advances the development of intelligent, transparent, and user-aligned heuristics for real-world optimization problems.
\end{abstract}


%
\IEEEpeerreviewmaketitle

\section{Introduction}
\IEEEPARstart{H}{euristics} are indispensable tools for solving complex decision-making and optimization problems, with applications spanning scheduling \cite{hu2025priority}, routing \cite{muriyatmoko2024heuristics}, and resource allocation \cite{gadi2025developing}. They are designed to provide adaptive, domain-specific solutions that balance solution quality and computational efficiency, enabling practitioners to make near-optimal decisions in real time. Among the diverse methodologies for heuristic design, Genetic Programming (GP) \cite{koza1994genetic} has emerged as a particularly powerful paradigm, capable of evolving interpretable symbolic rules that adapt to different problem structures \cite{zhong2018multifactorial}. GP-generated heuristics often rival, and sometimes surpass, learning-based methods such as neural combinatorial optimization \cite{mei2022explainable}, especially in terms of transparency and adaptability. Crucially, GP produces symbolic decision rules that can be inspected and understood by humans, in stark contrast to the opaque representations of most deep learning models.

Despite these advantages, the practical deployment of GP-evolved heuristics faces two persistent challenges: complexity and transferability. First, for large-scale real-world problems such as job shop scheduling, heuristics evolved by GP can become structurally intricate, with deeply nested expressions that obscure the underlying decision logic \cite{mei2022explainable}. While theoretically interpretable, such heuristics often remain inaccessible to human practitioners, undermining one of GP’s central promises. Second, GP typically evolves heuristics independently for each task, treating every new optimization scenario as a tabula rasa. This lack of knowledge reuse inflates computational costs and prevents the leveraging of prior evolutionary experience, reducing both efficiency and generalization in dynamic or multi-task environments. These challenges limit the practical adoption of GP, especially in high-stakes domains where interpretability, adaptability, and rapid deployment are critical.

In parallel, the rise of Large Language Models (LLMs) \cite{zhao2023survey}, such as ChatGPT \cite{OpenAI}, has opened transformative opportunities in artificial intelligence. LLMs have demonstrated extraordinary capabilities in understanding, reasoning with, and generating structured and unstructured information \cite{kocon2023chatgpt}. Beyond natural language, LLMs are increasingly being explored for symbolic reasoning, program synthesis, and scientific discovery \cite{wang2025large}, making them highly relevant for domains where both expressiveness and interpretability are required. In particular, their ability to summarize, abstract, and communicate complex patterns positions LLMs as promising tools for enhancing the interpretability and transferability of GP-evolved heuristics. By bridging symbolic decision rules with natural language explanation, LLMs can function as an interface between evolved heuristics and human practitioners, facilitating both comprehension and reuse of heuristic knowledge.

In this context, we propose \emph{EvoSpeak}, a novel framework that integrates LLMs with GP for interpretable and generalizable heuristic evolution. EvoSpeak leverages the unique strengths of both paradigms: GP’s ability to evolve powerful problem-specific heuristics, and LLMs’ capacity for abstraction, explanation, and knowledge transfer \cite{gupta2017insights}. Specifically, EvoSpeak positions LLMs as strategic partners that operate at the critical stages surrounding GP: seeding the evolutionary process with high-quality heuristics and elucidating the decision logic of the final heuristics. LLMs are employed to analyze GP-evolved heuristics, extract latent strategies and symbolic relationships, and re-express them in accessible forms. These insights can then be used to warm-start subsequent evolutionary runs, guide the search toward promising regions of the search space, and facilitate cross-task transfer. Additionally, LLMs are explored as generators of diverse initial populations aligned with user-specified preferences, thereby accelerating evolutionary search while incorporating domain knowledge.

This study focuses on scheduling as the primary application domain, reflecting its foundational importance in operations research and its broad impact on manufacturing \cite{giret2015sustainability}, logistics \cite{chen2024coordinated}, healthcare \cite{nabavizadeh2024mixed}, and project management \cite{levner2024fast}. Scheduling represents a longstanding and prominent application area for GP, where evolved heuristics offer problem-specific adaptability \cite{xu2023genetic}, yet often suffer from the aforementioned limitations of complexity and lack of transferability. By situating EvoSpeak in this challenging and practically significant domain, we aim to showcase both the feasibility and transformative potential of combining GP and LLMs.

We posit that LLMs contribute to heuristic evolution in at least three synergistic ways. Firstly, they enable \textbf{knowledge extraction} from evolved heuristics. GP heuristics, while symbolic, may encode sophisticated decision strategies that are not easily deciphered manually. LLMs can analyze these structures to extract underlying principles, symbolic motifs, and domain-relevant insights. Such extraction not only enhances interpretability but also informs subsequent evolutionary runs, providing a warm start that reduces computational cost and improves heuristic quality. Secondly, LLMs facilitate \textbf{knowledge transfer} \cite{lin2024multiobjective} across tasks. By abstracting and generalizing principles from one set of heuristics, LLMs can adapt them to related problem instances, reducing the need to evolve heuristics from scratch and enabling rapid adaptation to dynamic environments \cite{renke2021review}. This represents a step toward lifelong or continual learning in GP. Finally, LLMs enhance \textbf{interpretability} for practical adoption. In real-world applications, heuristic transparency is often a prerequisite for trust, compliance, and usability. LLMs, with their natural language generation capabilities, can decode and explain evolved heuristics in human-accessible formats, bridging the gap between symbolic evolution and practical deployment. They can also propose refinements or controlled variations, expanding the utility of evolved heuristics beyond their original context.

Overall, this work introduces EvoSpeak as a paradigm shift in heuristic evolution: moving from purely symbolic search toward an integrated framework where LLMs act as interpreters, teachers, and collaborators. The contributions of this study can be summarized as follows:
\begin{enumerate}
    \item We propose EvoSpeak, which learns from existing high-quality heuristics, extracts useful knowledge, and generates warm-start populations. By learning from knowledge, the framework accelerates evolutionary search and guides the discovery of more effective heuristics.
    \item We demonstrate that EvoSpeak can decode and summarize complex GP trees into natural-language explanations, thereby addressing the long-standing challenge of GP opacity and improving transparency and trust in the learned heuristics.
    \item We verify that EvoSpeak enables the transfer of heuristic knowledge across different problem instances and tasks. This reduces redundancy in evolution, improves adaptability to dynamic environments, and supports continual learning.
    \item We conduct extensive experiments on dynamic flexible job shop scheduling (DFJSS), under both single- and multi-objective formulations, and demonstrate that EvoSpeak yields more effective, preference-aware heuristics together with interpretable reports of their decision logic.
\end{enumerate}

By combining the evolutionary power of GP with the reasoning and interpretive capacities of LLMs, EvoSpeak introduces a new paradigm for heuristic discovery and deployment. While scheduling is employed as the primary case study, the principles underlying EvoSpeak are broadly applicable across domains where heuristics are indispensable. This integration not only improves efficiency, adaptability, and interpretability but also extends the frontier of evolutionary computation by situating LLMs as collaborators in symbolic search. 

The remainder of the paper is structured as follows: Section \ref{background} reviews related work on GP for heuristic evolution, LLMs, and interpretability challenges. Section \ref{method} presents the EvoSpeak framework in detail. Section \ref{design} outlines the experimental design, while Sections \ref{results} and \ref{further} report and discuss main results and empirical findings. Section \ref{conclusion} concludes with contributions, limitations, and avenues for future research.

\section{Background}
\label{background}

\subsection{Genetic Programming for Learning Heuristics}
GP has long been recognized as a powerful hyper-heuristic for the automated discovery of computer programs or decision-making rules to solve complex tasks \cite{koza1999genetic}. Unlike direct optimization approaches that focus on solving a single problem instance, GP operates at the meta-level by evolving heuristics—generalized rules or strategies—that can be deployed across diverse decision-making scenarios requiring rapid, real-time responses \cite{xu2023genetic2}. This ability to automatically evolve reusable and adaptive heuristics is particularly valuable in domains where manually designing effective strategies is infeasible due to high complexity, stochasticity, or dynamically changing environments.

Scheduling problems represent one of the most prominent and successful applications of GP. In particular, dynamic job shop scheduling has attracted extensive attention due to its inherent uncertainty and demand for online decision-making \cite{jakobovic2006dynamic, zhou2020automatic, chen2025optimizing, xu2025pareto}. Early research primarily targeted single-objective formulations, where GP was used to evolve algebraic priority rules based on shop-floor attributes (e.g., job waiting time, machine workload) with the aim of minimizing objectives such as mean flowtime or tardiness \cite{jakobovic2006dynamic}. Evaluated through discrete-event simulations, these GP-generated rules consistently outperformed handcrafted heuristics, establishing GP as a viable framework for capturing effective scheduling logic automatically in dynamic settings \cite{jakobovic2012evolving}.

As research matured, the inherently multi-criteria nature of real-world scheduling problems motivated the integration of multi-objective evolutionary algorithms (e.g., NSGA-II \cite{deb2002fast}, MOEA/D \cite{zhang2007moea}, SMS-EMOA \cite{xu2024preliminary}) into GP. These multi-objective extensions enabled the evolution of Pareto-optimal sets of heuristics balancing trade-offs among makespan, tardiness, energy consumption, and machine utilization \cite{zhang2019evolving, xu2023multi}. This transition from single-objective to multi-objective GP highlighted the capacity of evolutionary hyper-heuristics to accommodate conflicting requirements and to support decision-makers with sets of diverse yet effective solutions.

Beyond these core advancements, numerous innovations have been developed to improve GP’s scalability, robustness, and expressiveness. Grammar-guided and semantic GP constrain the search to syntactically and semantically meaningful spaces, improving both interpretability and efficiency \cite{whigham1995grammatically, vanneschi2014survey, xu2023semantic}. Surrogate-assisted GP addresses the high computational cost of simulation-based evaluation by approximating heuristic performance with predictive models \cite{gil2023surrogate}. Hybrid and ensemble-based approaches combine multiple complementary heuristics to enhance robustness across diverse operating conditions \cite{xu2023genetic2}. Together, these developments illustrate how GP has evolved from producing simple rules to generating rich heuristic portfolios that are powerful enough to address the complexities of real-world scheduling.

\begin{figure*}[t]
\centerline{\includegraphics[width=1.0\textwidth]{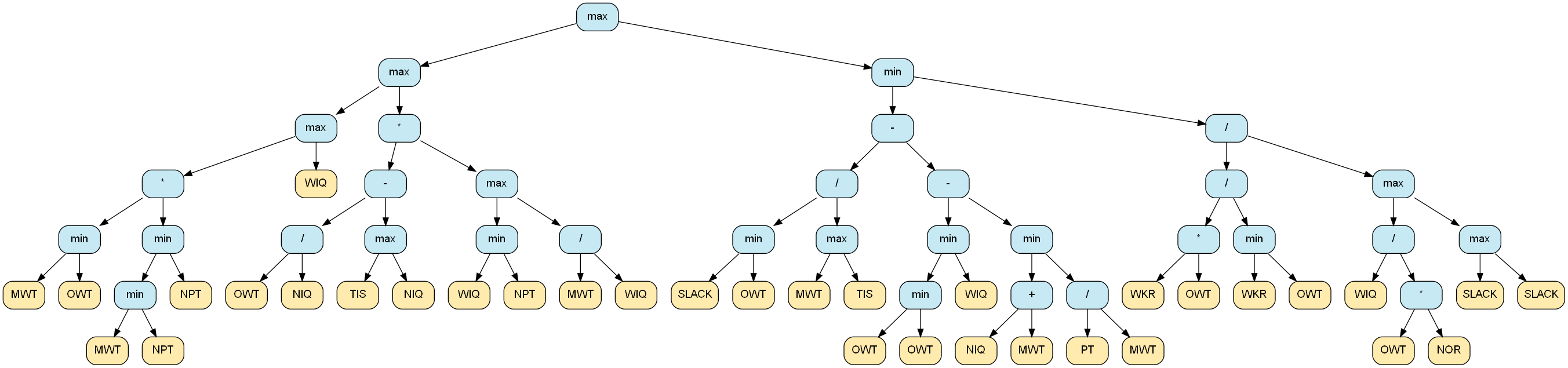}}
\caption{An example of a tree-structured routing rule evolved for a scheduling problem. Yellow nodes denote system features (e.g., WIQ, MWT), while blue nodes denote operators (e.g., max, min, subtraction). While effective, such trees can be large and difficult to interpret.}
\label{fig: tree-structure}
\end{figure*}

\subsection{Interpretability \& Natural Forms in Heuristics}
Although GP heuristics are represented as symbolic trees and are therefore, in principle, interpretable, in practice they often grow into large, deeply nested structures whose decision logic is difficult for humans to follow \cite{mei2022explainable}, as illustrated in Fig.~\ref{fig: tree-structure}. This phenomenon, known as \emph{bloat}, results in oversized trees that obscure heuristic logic and reduce interpretability.

Several bloat control techniques have been proposed to mitigate this issue \cite{luke2006comparison, silva2009dynamic}. For, example, parsimony pressure \cite{zhang1995balancing, poli2013parsimony} penalizes overly large individuals during fitness evaluation, while depth limits and adaptive size constraints impose direct bounds on tree growth. Yet, strong size control may reduce population diversity or prematurely discard useful building blocks, creating a fundamental trade-off between simplicity and performance. To further enhance interpretability, grammar-guided GP introduces syntactic and semantic restrictions that bias evolution toward human-readable or domain-relevant structures \cite{ingelse2025comparison, montana1995strongly}. These approaches encourage the generation of heuristics in the form of algebraic rules or domain-specific functional templates that practitioners can more easily validate. Recent advances also demonstrate how embedding principles such as symmetry, invariance, and domain knowledge can guide the search toward compact yet expressive symbolic representations \cite{liu2023learning, crabbe2023evaluating, bhattarai2023interpretable}.

In scheduling, these principles translate into heuristics that are not only effective but also transparent and actionable. Simple priority indices or weighted linear combinations of shop-floor attributes, for instance, are much easier for practitioners to adopt and trust compared with deeply nested expressions. However, despite progress with parsimony, grammar constraints, and symbolic regression, existing methods remain limited in their ability to \emph{summarize}, \emph{contextualize}, or \emph{communicate} the deeper logic embedded in large GP trees. This persistent gap motivates the introduction of LLMs, which offer the capability to translate symbolic structures into natural language explanations, identify semantic patterns, and even propose simplified alternatives. Such integration shifts interpretability from structural constraints alone toward an active process of human-centered communication.

\subsection{LLMs in Evolutionary Computation}
LLMs, such as ChatGPT \cite{OpenAI}, represent a significant advancement in artificial intelligence. Trained on massive and diverse corpora, these models exhibit strong capabilities in understanding, generating, and reasoning about both structured and unstructured data \cite{chang2024survey}. Their aptitude for processing symbolic representations, identifying latent patterns, and contextualizing information makes them particularly well-suited for tasks that demand nuanced reasoning and complex decision-making.

In recent years, there has been growing interest in integrating LLMs into evolutionary computation (EC) frameworks to improve heuristic design and evolution \cite{liu2024evolution, zhang2024understanding, ye2024reevo, yao2024multi}. This emerging research direction typically employs one or more of the following paradigms: 1) LLMs-as-Generator \cite{hemberg2024evolving, romera2024mathematical, zhang2025llm} — The LLMs produce code fragments or natural-language ``thoughts'' that represent candidate heuristics, with EC subsequently filtering, selecting, and refining them. 2) LLMs-as-Evolutionary Operator \cite{guo2023connecting, liu2024large, liu2024language} — The LLMs replace traditional hand-coded variation operators (e.g., mutation, crossover) by generating candidate variants conditioned on one or more parent solutions. 3) LLMs-as-Evaluator/Explainer \cite{maddigan2024explaining} — The LLMs analyze, interpret, or evaluate candidate solutions (e.g., GP trees) to improve interpretability, debugging, and semantic guidance. 4) Hybrid LLMs+EC \cite{ye2024reevo, qi2025memetic, shi2025generalizable} — The LLMs generate diverse candidates, while EC performs population-based search and local refinements; some approaches add reflection or memory mechanisms to iteratively improve reasoning and performance. 

Several representative studies illustrate these approaches. Liu et al. \cite{liu2024evolution} proposed EoH, where heuristics are represented as natural-language ``thoughts'' generated by LLMs and evolved using EC to exploit LLMs creativity while retaining the exploratory power of population-based search. Hemberg et al. \cite{hemberg2024evolving} formalized an LLM-driven evolutionary algorithm that directly integrates LLMs into variation operators for evolving code-like solutions, providing both algorithmic design details and empirical comparisons. Zhang et al. \cite{zhang2024understanding} applied LLM-based semantic analysis to preserve meaningful code structures during mutation. Ye et al. \cite{ye2024reevo} introduced ReEvo, where LLMs suggest context-aware modifications to evolved heuristics through reflective guidance, while Yao et al. \cite{yao2024multi} explored multi-agent LLMs frameworks for collaborative heuristic development. Collectively, these studies highlight the potential of LLMs to act as creative engines within EC, generating novel heuristics, guiding variation operators, and enabling co-evolutionary search strategies.

Despite these promising developments, several limitations remain. First, most work has been evaluated on small-scale or relatively simple domains, where heuristic search is computationally less demanding. Population sizes are typically limited to around 20 individuals due to the high cost of frequent LLM queries, which severely restricts scalability to large or complex search spaces. Second, the reliance on LLM calls for every crossover or mutation step imposes substantial computational overhead, making these methods impractical for time-sensitive or large-scale optimization. Finally, current methods have largely been tested on static problem instances, leaving their robustness in dynamic environments—such as the DFJSS problem, where tasks and resources change over time—largely unexplored. The large-scale, dynamic nature of DFJSS demands heuristic design approaches that are computationally efficient, scalable, adaptable, and maintain interpretability and transferability. Heavy reliance on direct LLMs integration within the evolutionary loop fails to meet these requirements due to prohibitive runtime costs and limited scalability. Together, these issues render the direct, online integration of LLMs into the evolutionary process impractical for complex scheduling domains.

To overcome these issues, our work introduces EvoSpeak, a paradigm where LLMs are positioned not as embedded operators but as offline enablers before and after GP. LLMs are used before GP to extract knowledge from existing heuristics and generate warm-start populations aligned with user preferences, and after GP to interpret evolved heuristics into human-readable reports. This design eliminates costly per-iteration queries, retains GP’s scalability in large dynamic environments, and simultaneously addresses the enduring challenges of efficiency, transferability, and interpretability in heuristic evolution.

\subsection{Dynamic Flexible Job Shop Scheduling}
DFJSS is a representative and practically important decision-making problem in manufacturing \cite{nie2013gep}. Unlike static job shop scheduling, DFJSS considers dynamic shop-floor environments in which disruptions such as new job arrivals and priority changes occur unpredictably \cite{ren2022joint}. The objective is to design scheduling heuristics that not only deliver high-quality schedules but also remain adaptive, interpretable, and responsive to diverse and evolving operational goals.  

Formally, let $\mathcal{J} = \{J_1, J_2, \dots, J_n\}$ denote the set of jobs. Each job $J_i$ is released at time $r_i$, has a weight $\rho_i$, and a due date $d_i$. A job is composed of an ordered sequence of operations $\mathcal{O}_i = \{O_{i1}, O_{i2}, \dots, O_{im_i}\}$, where $m_i$ is the number of operations. Each operation $O_{ik}$ can be processed on a subset of machines $\mathcal{M}_{ik} \subseteq \mathcal{M} = \{M_1, M_2, \dots, M_h\}$, with machine-dependent processing times $p_{ik}^{(j)}$ on $M_j \in \mathcal{M}_{ik}$. If operations are executed on geographically distributed machines, additional transportation delays $\tau_{k_1,k_2}$ are incurred. The scheduling process is subject to the following constraints:

\begin{itemize}
    \item \textbf{Operation precedence.}  
    Operations within the same job must follow their predefined order:
    \begin{equation}
        s_{i,k+1} \geq c_{ik}, \quad \forall i, k,
    \end{equation}
    where $s_{ik}$ and $c_{ik}$ denote the start and completion times of operation $O_{ik}$.

    \item \textbf{Non-preemption.}  
    Once started, an operation must run to completion without interruption:
    \begin{equation}
        c_{ik} = s_{ik} + p_{ik}^{(j)}, \quad \forall i,k, \; M_j \in \mathcal{M}_{ik}.
    \end{equation}

    \item \textbf{Machine capacity.}  
    Each machine can process at most one operation at any given time:
    \begin{equation}
        \sum_{i,k} \mathbb{1}\{O_{ik} \text{ is processed on } M_j \text{ at time } t\} \leq 1, 
        \quad \forall t, M_j,
    \end{equation}
    where $\mathbb{1}\{\cdot\}$ is the indicator function.

    \item \textbf{Machine assignment.}  
    Each operation must be allocated to exactly one eligible machine:
    \begin{equation}
        O_{ik} \mapsto M_j, \quad M_j \in \mathcal{M}_{ik}.
    \end{equation}
\end{itemize}

The optimization objectives in DFJSS vary according to production requirements. In this work, we consider five widely studied performance measures:  

\begin{itemize}
    \item \textbf{Max tardiness.}  
    The tardiness of job $J_i$ is defined as $T_i = \max\{0, c_{im_i} - d_i\}$, where $c_{im_i}$ is its completion time and $d_i$ its due date. Minimizing the maximum tardiness improves robustness by reducing the worst-case violation of due dates:
    \begin{equation}
        T_{max} = \max_{i=1,\dots,n} T_i.
    \end{equation}

    \item \textbf{Mean tardiness.}  
    The mean tardiness assesses the average delay of jobs beyond their due dates:
    \begin{equation}
        T_{mean} = \frac{1}{n} \sum_{i=1}^n T_i.
    \end{equation}

    \item \textbf{Mean flowtime.}  
    The flowtime of job $J_i$ is $F_i = c_{im_i} - r_i$, where $r_i$ is its release time. The mean flowtime evaluates the average time jobs spend in the system, reflecting overall responsiveness:
    \begin{equation}
        F_{mean} = \frac{1}{n} \sum_{i=1}^n F_i.
    \end{equation}

    \item \textbf{Mean weighted tardiness.}  
    To account for job importance, weights $\rho_i$ are introduced, yielding the mean weighted tardiness:
    \begin{equation}
        WT_{mean} = \frac{1}{n} \sum_{i=1}^n \rho_i T_i,
    \end{equation}
    which balances timely delivery with job priorities.

    \item \textbf{Mean weighted flowtime.}  
    Similarly, incorporating weights into flowtime yields the mean weighted flowtime:
    \begin{equation}
        WF_{mean} = \frac{1}{n} \sum_{i=1}^n \rho_i F_i,
    \end{equation}
    which captures both system efficiency and priority-aware responsiveness.
\end{itemize}

Depending on user preferences, these objectives may be optimized individually or in combination. A weighted-sum formulation provides a general framework:  
\begin{equation}
    \min_{\pi \in \Pi} \; F(\pi) = \sum_{j=1}^{m} \lambda_j f_j(\pi),
\end{equation}
where $\pi$ denotes a scheduling heuristic, $\Pi$ is the feasible heuristic space, $m$ is the number of objectives, $f_j \in \{T_{max}, T_{mean}, WT_{mean}, F_{mean}, WF_{mean}\}$ represents the $j$-th objective, and $\lambda_j \in [0,1]$ is the corresponding user-specified preference, subject to $\sum_{j=1}^{m} \lambda_j = 1$.

Overall, DFJSS is a particularly challenging problem because it requires scheduling heuristics that can produce high-quality solutions in real time under uncertain and dynamic shop-floor conditions. Disruptions such as unexpected job arrivals or priority changes demand heuristics that are both computationally efficient and highly adaptable. GP has been extensively applied to DFJSS to evolve symbolic tree-based heuristics that map system states to scheduling decisions \cite{guo2024improved}. Enhancements including surrogate-assisted GP \cite{zhou2020automatic}, ensemble GP \cite{xu2023genetic2}, and multi-objective GP \cite{xu2023multi} have improved responsiveness, robustness, and multi-objective handling. However, interpretability remains a key bottleneck: the complex heuristics produced by GP are often difficult to analyze or communicate, limiting their trustworthiness and practical usability. At the same time, DFJSS is inherently multiobjective, with stakeholders potentially prioritizing flowtime, tardiness, or priority satisfaction. Balancing these objectives while incorporating user preferences further increases the problem complexity. These challenges highlight the limitations of conventional GP-based heuristic design, which can generate effective rules but often in opaque forms that are hard to interpret or transfer. To address this gap, we propose a novel integration of GP with LLMs. GP serves as the evolutionary engine for discovering high-performing heuristics, while LLMs provide a reasoning and interpretability layer.

\section{EvoSpeak}
\label{method}

\begin{figure*}[t]
\centerline{\includegraphics[width=1.0\textwidth]{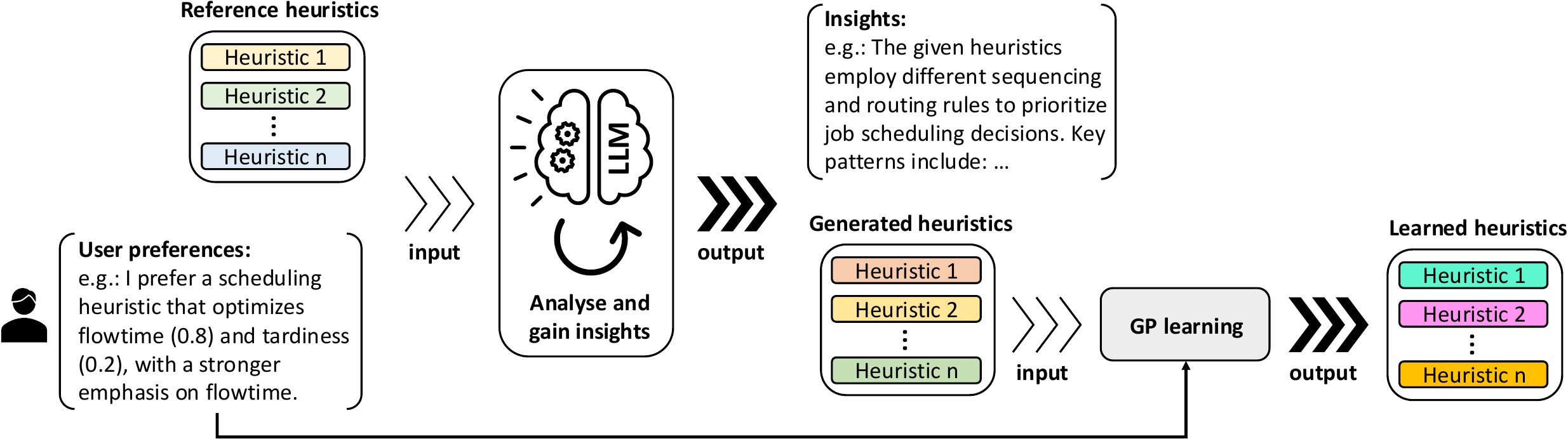}}
\caption{Overall EvoSpeak framework. LLMs act as both a knowledge extraction engine and a symbolic interpreter, integrated into the GP loop.}
\label{fig: overallframework}
\end{figure*}

We introduce EvoSpeak, an LLM-assisted GP framework designed to accelerate the evolution of heuristics, improve their transferability across tasks, and enhance their interpretability. Unlike approaches that embed LLM queries directly into the evolutionary loop, EvoSpeak strategically employs LLMs before and after the GP process: prior to evolution, LLMs extract knowledge from existing heuristics and synthesize informed initial populations; after evolution, they translate evolved heuristics into human-readable explanations. This design avoids scalability bottlenecks while fully exploiting the symbolic reasoning and generative capabilities of LLMs. The overall architecture is illustrated in Fig.~\ref{fig: overallframework} and summarized in Algorithm~\ref{alg:EvoSpeak}.

\subsection{Main Framework}
Let $\mathcal{H} = {h_1, h_2, \dots, h_C}$ denote a reference set of heuristics, where each $h_i: \mathcal{S} \to \mathbb{R}$ maps a scheduling state $s \in \mathcal{S}$ to a priority score. EvoSpeak introduces an LLM-driven meta-mapping function:
\begin{equation}
\mathcal{M}_{\mathrm{LLM}}: (\mathcal{H}, \Lambda) \rightarrow \mathcal{H}',
\end{equation}
where $\Lambda$ encodes user preferences (e.g., weights for multiple objectives) and $\mathcal{H}'$ denotes a synthesized set of heuristics aligned with those preferences.

The workflow consists of four integrated stages:
\begin{itemize}
    \item \textbf{Heuristic Collection.} Candidate heuristics are obtained either by running GP on a representative task instance $\gamma \in \Gamma$ or by importing heuristics evolved in prior studies. This library forms the raw material for knowledge extraction.
    \item \textbf{Population Initialization via LLMs.} EvoSpeak applies the LLM to analyze $\mathcal{H}$, extracting symbolic motifs, recurrent structural patterns, and implicit decision-making rules. By combining this extracted knowledge with $\Lambda$, the system generates a knowledge-rich initial population $\mathcal{P}_0 = \mathcal{M}_{\mathrm{LLM}}(\mathcal{H}, \Lambda)$ that provides GP with a strong warm start.
    \item \textbf{Heuristic Evolution.} GP refines the population over $G$ generations using standard operators ${\mathsf{mutation}, \mathsf{crossover}}$ guided by a weighted fitness function:
    \begin{equation}
    F(h) = \sum_{k=1}^K w_k f_k(h),
    \end{equation}
    where $f_k$ represents the $k$-th objective (e.g., mean flowtime, tardiness), and $w_k$ its weight. Starting from $\mathcal{P}_0$, the search focuses on promising subspaces of heuristics, accelerating convergence.
    \item \textbf{Interpretability Enhancement.} The final best heuristic $h^\ast$ is passed back to the LLM for symbolic-to-natural-language translation:
\begin{equation}
R_{\mathrm{explain}} = \mathcal{M}_{\mathrm{LLM}}^{\mathrm{explain}}(h^\ast),
\end{equation}
producing structured, human-readable explanations of the decision-making logic. These explanations support both expert analysis and stakeholder communication.
\end{itemize}

This integration allows EvoSpeak to retain the scalability of GP while injecting LLM-guided knowledge at critical pre- and post-evolution stages, improving efficiency, transferability, and interpretability.

\begin{algorithm}[t]
    \caption{EvoSpeak Algorithm}
    \footnotesize
    \label{alg:EvoSpeak}
    \SetAlgoLined
    \KwIn{Population size $N$, Generations $G$, Reference heuristics $\mathcal{H}$, Problem set $\Gamma$, LLM $\mathcal{M}_{\mathrm{LLM}}$, Preferences $\Lambda$}
    \KwOut{Best heuristic $h^\ast$}
    $\mathcal{P}_0 \leftarrow \mathcal{M}_{\mathrm{LLM}}(\mathcal{H}, \Lambda)$ \tcp*{LLM-based initialization}
    \For{$t \gets 0$ \KwTo $G-1$}{
        Evaluate $F(h)$ for all $h \in \mathcal{P}_t$\;
        Select parents via tournament selection\;
        Apply crossover and mutation to produce $\mathcal{P}_{t+1}$\;
        $h^\ast \gets \arg\max_{h \in \mathcal{P}_t} F(h)$\;
    }
    $R_{\mathrm{explain}} \leftarrow \mathcal{M}_{\mathrm{LLM}}^{\mathrm{explain}}(h^\ast)$ \tcp*{LLM interpretability}
    \Return{$h^\ast$}
\end{algorithm}

\subsection{Knowledge Extraction and Population Initialization}
\label{initialisation}
The initialization phase is central to EvoSpeak. Given $\mathcal{H} = {h_1, \dots, h_C}$, the LLM serves as a knowledge extractor:
\begin{equation}
\mathcal{K}_{\mathrm{LLM}}: \mathcal{H} \longrightarrow \mathcal{Z},
\end{equation}
where $\mathcal{Z}$ encodes latent decision principles such as symbolic invariants, operator preferences, and performance-sensitive structures.

The LLM then synthesizes the initial population:
\begin{equation}
\mathcal{P}_0 = \mathcal{M}_{\mathrm{LLM}}(\mathcal{Z}, \Lambda),
\end{equation}
where $\Lambda$ specifies the objective trade-offs. Unlike random initialization, $\mathcal{P}_0$ incorporates historical experience and user intent, yielding a population predisposed to good performance and faster convergence.

\begin{figure}[t]
\centerline{\includegraphics[width=0.5\textwidth]{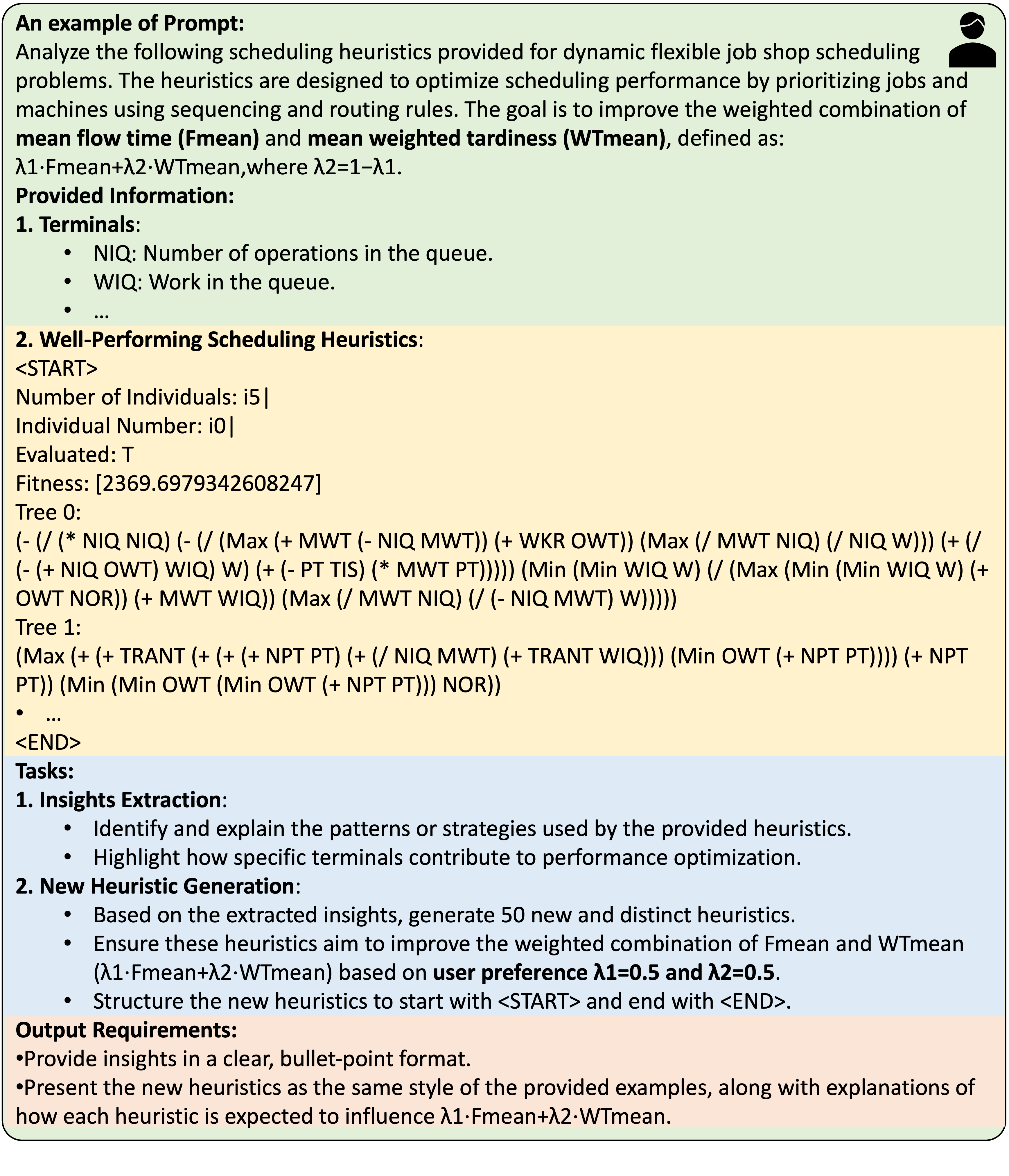}}
\caption{An example prompt for population initialization using an LLM, incorporating existing scheduling heuristics and user preferences for a multi-objective DFJSS problem.}
\label{fig: prompt}
\end{figure}

Prompt design plays a critical role. EvoSpeak prompts the LLM for $\mathcal{M}_{\mathrm{LLM}}$ with (i) problem context, (ii) reference heuristics and terminal semantics, (iii) task-specific preferences, and (iv) explicit output constraints. Fig.~\ref{fig: prompt} shows an example where the LLM is asked to analyze heuristics for a DFJSS problem and generate variants that balance mean flowtime and tardiness. In this example, we first define the overall requirement, specifying that the LLM should analyze reference heuristics and optimize a weighted combination of two objectives: mean flowtime and mean weighted tardiness. To ensure clarity, we explicitly provide the equation for the objective function. Additionally, we include explanations of how the reference heuristics operate, along with details about their structures and the meanings of the terminal symbols used. Next, we clearly outline the tasks assigned to the LLM. Specifically, we request insight extraction from the provided heuristics, followed by the generation of new heuristics based on the extracted knowledge. Finally, we specify the required output format to ensure clarity, consistency, and usability. By carefully structuring the prompt, EvoSpeak ensures that the generated heuristics are both interpretable and directly usable by GP.

\subsection{Knowledge Transfer Across Tasks}
\label{sec:transfer}
EvoSpeak extends beyond single-task optimization by enabling cross-task generalization. Suppose $\Gamma_{\mathrm{src}}$ and $\Gamma_{\mathrm{tgt}}$ denote source and target task distributions. Knowledge $\mathcal{Z}_{\mathrm{src}} = \mathcal{K}_{\mathrm{LLM}}(\mathcal{H}_{\mathrm{src}})$ extracted from source heuristics is adapted to bootstrap evolution on $\Gamma_{\mathrm{tgt}}$:
\begin{equation}
\mathcal{P}_0^{\mathrm{tgt}} = \mathcal{M}_{\mathrm{LLM}}\big(\mathcal{T}_{\mathrm{adapt}}(\mathcal{Z}_{\mathrm{src}}, \Gamma_{\mathrm{tgt}}), \Lambda_{\mathrm{tgt}}\big).
\end{equation}
where $\mathcal{T}_{\mathrm{adapt}}$ performs knowledge adaptation through:
\begin{enumerate}
    \item Principle Generalization: Extracting task-invariant components $I(h)$ that preserve core decision-making logic.
    \item Instance Adaptation: Refining parameters using target-specific statistics $\Phi(\Gamma_{\mathrm{tgt}})$.
    \item Cross-Domain Mapping: Translating heuristics between distinct state or feature spaces.
\end{enumerate}

This mechanism prevents GP from ``starting over'' for every task, instead enabling it to fine-tune pre-adapted populations. For example, a heuristic optimized for scheduling in a low-load manufacturing plant can be adapted, with minimal modifications, to efficiently handle a high-load environment. As a result, EvoSpeak reduces training costs and accelerates deployment in dynamic, heterogeneous environments.

\subsection{Interpretability Enhancement}
\label{reportLLM}
Although GP is capable of evolving highly effective heuristics, the resulting symbolic expressions are often complex and difficult for practitioners to interpret. For example:
\begin{equation}
    h(s) = \frac{p_1(s) + p_2(s) \cdot \exp(\mathrm{PT} / \mathrm{TIS})}{\max\{1, p_3(s)\} + \log(1 + p_4(s))},
\end{equation}
where $p_i(s)$ are symbolic sub-expressions (sub-trees), PT denotes processing time, and TIS represents time in system. While such heuristics may achieve strong performance, their underlying decision rationale can remain opaque to human users.

To bridge this gap, EvoSpeak employs LLMs as symbolic interpreters, mapping expressions into structured, natural language explanations, thereby enhancing interpretability and fostering user trust. Building upon the initialization framework in Section~\ref{initialisation}, we extend the LLM’s role to generate a structured, user-friendly analysis report. This report organizes technical explanations into clear narratives, providing detailed reasoning for expert users while offering concise, accessible summaries for non-experts. The interpretability mapping is defined as:
\begin{equation}
    R_{\mathrm{explain}} = \mathcal{M}_{\mathrm{LLM}}^{\mathrm{explain}}(h),
\end{equation}
where $\mathcal{M}_{\mathrm{LLM}}^{\mathrm{explain}}$ translates $h$ into domain-relevant, plain-language descriptions, e.g.,
\begin{equation}
\notag
\begin{matrix}
\text{\emph{``PT plays a more important role in scenarios where}}\\
\text{\emph{optimizing flowtime is the primary objective...''}}
\end{matrix}
\end{equation}
This explanation framework ensures that each heuristic is both \emph{performant} and \emph{transparent}, a critical requirement in high-stakes domains such as healthcare, manufacturing, and other safety-critical decision-making environments.

By making evolved heuristics transparent and narratively interpretable, EvoSpeak supports user trust and facilitates adoption in safety-critical contexts such as healthcare scheduling, manufacturing, and logistics.

\section{Experiment Design}
\label{design}
\subsection{Dataset} 
The experiments are conducted on the DFJSS simulation model introduced in \cite{xu2023genetic}, which provides a well-established yet challenging platform for evaluating heuristic scheduling methods. Each instance simulates 5,000 jobs—including a 1,000-job warm-up period—processed across 10 heterogeneous machines with randomly generated processing rates in the interval $[10,15]$. Transportation is explicitly modeled: machine-to-entry/exit distances are sampled from a discrete uniform distribution between 35 and 500 units, with transport speed fixed at 5 units. Job arrivals follow a Poisson process. Jobs are composed of a random number of operations, uniformly sampled from ${2,\dots,10}$, with workloads drawn from a discrete uniform distribution over $[100,1000]$. Job importance is represented through weights, with 20\% of jobs assigned weight 1, 60\% assigned weight 2, and 20\% assigned weight 4. Due dates are generated by adding $1.5$ times the total processing time to each job’s arrival time, creating instances with realistic congestion and tardiness pressures.

A central factor in DFJSS is shop utilization, which directly controls the degree of congestion and thus the difficulty of the scheduling task. To examine EvoSpeak under diverse operating conditions, we construct six single-objective and four multi-objective scenarios by varying utilization levels (0.85 and 0.95) and objective formulations. The performance measures include maximum tardiness (Tmax), mean flowtime (Fmean), mean tardiness (Tmean), and mean weighted tardiness (WTmean). In single-objective scenarios, only one of these measures is optimized. In multi-objective scenarios, trade-offs between two objectives are considered, with relative importance encoded by weights $\lambda_1$ and $\lambda_2=1-\lambda_1$. Scenarios are denoted by their objective(s) and utilization level, e.g., $\langle$Tmax, 0.85$\rangle$. The following cases are studied:
\begin{itemize}
    \item Scenarios \textless{}Tmax, 0.85\textgreater{} and \textless{}Tmax, 0.95\textgreater{}: Minimize Tmax at utilization levels of 0.85 and 0.95, respectively;
    \item Scenarios \textless{}Fmean, 0.85\textgreater{} and \textless{}Fmean, 0.95\textgreater{}: Minimize Fmean at utilization levels of 0.85 and 0.95, respectively;
    \item Scenarios \textless{}WTmean, 0.85\textgreater{} and \textless{}WTmean, 0.95\textgreater{}: Minimize WTmean at utilization levels of 0.85 and 0.95, respectively;
    \item Scenarios \textless{}Fmean-WTmean, 0.85\textgreater{} and \textless{}Fmean-WTmean, 0.95\textgreater{}: Minimize Fmean and WTmean at utilization levels of 0.85 and 0.95, respectively;
    \item Scenarios \textless{}Tmean-WFmean, 0.85\textgreater{} and \textless{}Tmean-WFmean, 0.95\textgreater{}: Minimize Fmean and WTmean at utilization levels of 0.85 and 0.95, respectively.
\end{itemize}

For each scenario, the evaluation uses 50 instances for training and a separate set of 30 unseen instances for testing.

\subsection{Parameter Setting}
\label{sec:parameters}
The proposed EvoSpeak framework builds on a symbolic GP backbone, extended with LLM-based initialization and interpretability modules. Table~\ref{notation} defines the terminal set, which incorporates features at four levels: machine-related (e.g., NIQ, WIQ, MWT), operation-related (e.g., PT, NPT, OWT), job-related (e.g., WKR, NOR, rDD, SLACK, W, TIS), and transportation-related (TRANT). These variables capture the local and global dynamics of DFJSS. The function set consists of standard arithmetic operators ($+,-,\times,/$) and min/max functions. Division is protected by returning $1$ when the denominator is zero, and min/max return the smaller or larger input, respectively. This design ensures closure and numerical stability of evolved heuristics.

GP parameters are summarized in Table~\ref{parameter}. Populations of 100 individuals are evolved for 50 generations, initialized using the ramped half-and-half method. A maximum tree depth of 8 is enforced to control code bloat. Genetic operators are configured with crossover, mutation, and reproduction rates of $0.80$, $0.15$, and $0.05$, respectively, with tournament selection (size 4) for parent choice. The EvoSpeak variant employs \texttt{ChatGPT~4.0} for both heuristic generation (population initialization and knowledge transfer) and heuristic interpretation. This configuration ensures that baseline GP and EvoSpeak are compared under identical evolutionary conditions, isolating the contributions of LLM-based knowledge extraction, transfer, and interpretability.

\begin{table}[t]
\caption{Terminal and function sets used in the GP framework for DFJSS.}
\label{notation}
\centering
\footnotesize
\begin{threeparttable}
\begin{tabular}{c|l}
\hline
\textbf{Notation} & \textbf{Description} \\ 
\hline
NIQ    & Number of operations in the queue \\
WIQ    & Work in the queue \\
MWT    & Machine waiting time $= t^* - \text{MRT}^*$ \\
PT     & Processing time of the operation \\
NPT    & Median processing time of the next operation \\
OWT    & Operation waiting time $= t - \text{ORT}^*$ \\
WKR    & Work remaining \\
NOR    & Number of operations remaining \\
rDD    & Relative due date $= \text{DD}^* - t$ \\
SLACK  & Job slack time \\
W      & Job weight \\
TIS    & Time in system $= t - \text{releaseTime}^*$ \\
TRANT  & Transportation time \\ 
\hline
Function & $+, -, \times, /, \max, \min$ \\
\hline
\end{tabular}
\begin{tablenotes}
\item[*] $t$: current time; MRT: machine ready time; DD: due date; ORT: operation ready time; releaseTime: release time.
\end{tablenotes}
\end{threeparttable}
\end{table}

\begin{table}[t]
\caption{Parameter settings of the GP methods.}
\label{parameter}
\centering
\footnotesize
\begin{tabular}{l|c}
\hline
\textbf{Parameter} & \textbf{Value} \\ \hline
Population size & $100$ \\
Generations & $50$ \\
Initial min/max depth & $2$ / $6$ \\
Maximal tree depth & $8$ \\
Terminal / non-terminal selection rate & $0.10$ / $0.90$ \\
Crossover / mutation / reproduction rate & $0.80$ / $0.15$ / $0.05$ \\
Tournament size & $4$ \\
LLM model used & \texttt{ChatGPT~4.0} \\
\hline
\end{tabular}
\end{table}

\subsection{Comparison Design}
To evaluate EvoSpeak, we design experiments that systematically examine how LLMs enhance heuristic evolution and interpretation compared to conventional GP. The study focuses on four complementary dimensions.
\begin{itemize}
    \item First, we assess whether LLMs can meaningfully extract knowledge from existing heuristics. EvoSpeak prompts the LLM with evolved heuristics and domain context, then evaluates whether the LLM-derived initial population outperforms a randomly initialized GP population. This measures the LLM’s ability to distill and reuse symbolic scheduling knowledge.
    \item Second, we examine knowledge transfer and warm starting. Here, heuristics evolved in source scenarios are summarized by the LLM and adapted to initialize GP in target scenarios. By comparing EvoSpeak against baseline GP (without transfer), we test whether LLM-assisted warm starts accelerate convergence and improve heuristic quality across unseen tasks.
    \item Third, we evaluate adaptability to user preferences in multi-objective scheduling. EvoSpeak leverages LLM prompting to adjust heuristic recommendations to different weightings between objectives (e.g., balancing Fmean vs. WTmean). This tests whether the framework can dynamically generate heuristics tailored to stakeholder-defined trade-offs.
    \item Finally, we investigate interpretability enhancement. Raw GP heuristics are passed to the LLM for translation into structured, human-readable explanations. These explanations are then qualitatively assessed for clarity, fidelity to the underlying logic, and usefulness in real-world scheduling contexts.
\end{itemize}

Together, these experiments rigorously benchmark EvoSpeak against traditional GP, highlighting not only improvements in optimization performance but also advances in adaptability and interpretability—qualities essential for the practical deployment of evolved scheduling heuristics.

\section{Experimental Results}
\label{results}

\subsection{Single-Objective Scenarios}

\subsubsection{Initial Population Performance Distribution}

\begin{figure}[t]
\centerline{\includegraphics[width=0.5\textwidth]{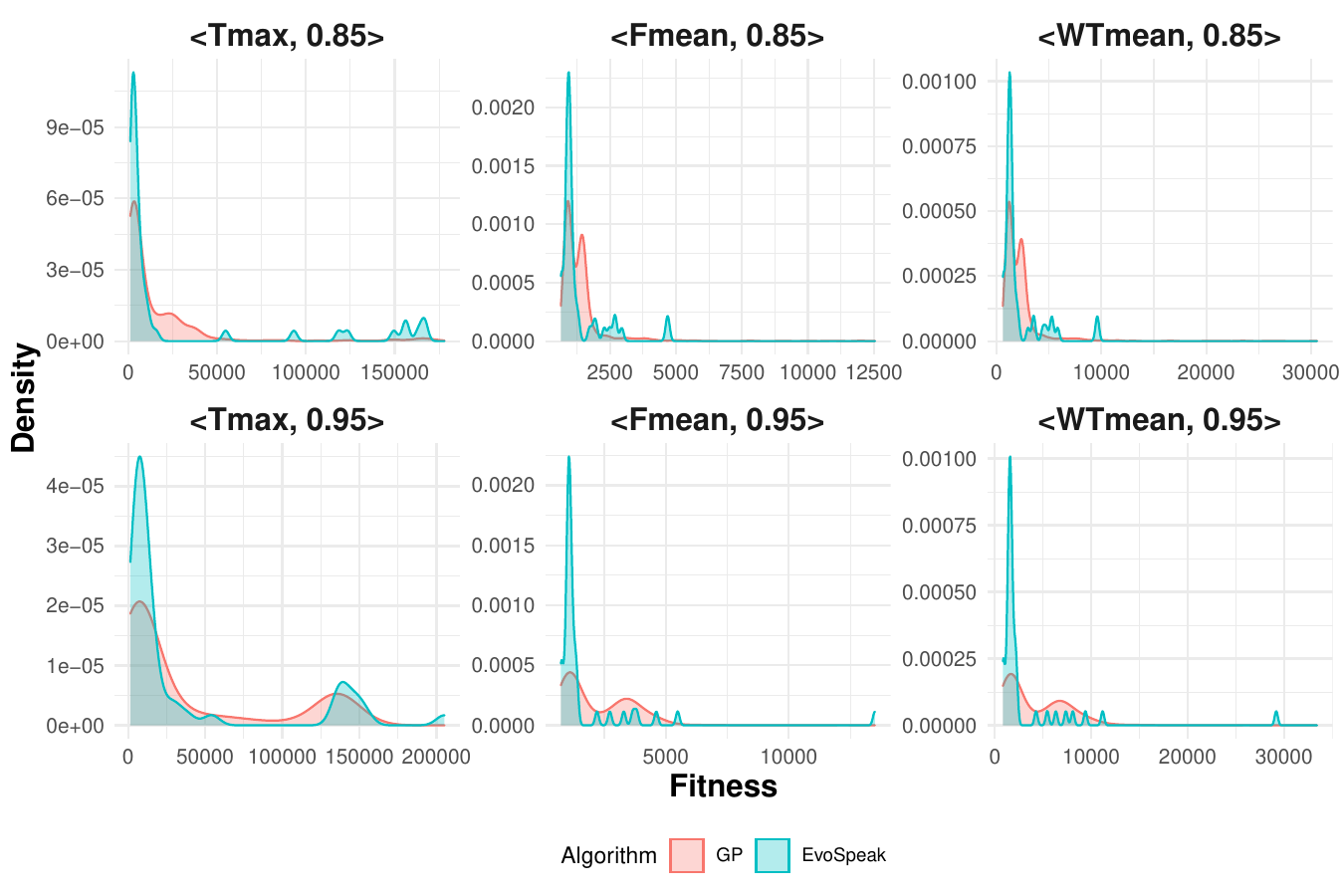}}
\caption{The fitness distribution of the initial population by GP and EvoSpeak.}
\label{fig: distribution}
\end{figure}

A key goal of EvoSpeak is to generate a more effective initial population of scheduling heuristics by analyzing existing heuristics, extracting structural knowledge, and producing novel candidates. Fig.~\ref{fig: distribution} compares the fitness density distributions of initial populations generated by standard GP and EvoSpeak across different scenarios. 

Peaks located further left indicate that a larger fraction of individuals exhibit lower (better) fitness values. As observed, in most scenarios, EvoSpeak (blue) has a peak farther left or overlaps with GP (red), indicating that EvoSpeak tends to find better heuristics more frequently. However, in some cases, EvoSpeak has a wider spread, suggesting higher variance, which implies that some generated scheduling heuristics may be less effective. On \textless{}Tmax, 0.85\textgreater{} and \textless{}Tmax, 0.95\textgreater{}, EvoSpeak shows a higher density near lower fitness values, suggesting better performance. On \textless{}Fmean, 0.85\textgreater{} and \textless{}Fmean, 0.95\textgreater{}, both methods exhibit similar distributions, but EvoSpeak has a greater density at lower fitness values. On \textless{}WTmean, 0.85\textgreater{} and \textless{}WTmean, 0.95\textgreater{}, EvoSpeak demonstrates a slightly broader spread but remains concentrated in the lower fitness range. While EvoSpeak occasionally produces broader distributions, reflecting higher variance, the presence of highly effective individuals enriches the initial search space and enhances evolutionary potential. Overall, EvoSpeak’s knowledge-guided initialization provides both quality and diversity, seeding GP with heuristics that accelerate convergence and improve final performance.

\subsubsection{Test Performance}
We further analyse the final mean and standard deviation test performance of 30 runs of both GP and EvoSpeak across six DFJSS scenarios. The uparrow ($\uparrow$) denotes statistically significant improvement by EvoSpeak over GP on that scenario under the Wilcoxon test (significant level = $0.05$) \cite{cuzick1985wilcoxon} as shown in Table \ref{table: test}.

\begin{table}[t]
\caption{The mean and standard deviation of test performance of GP and EvoSpeak by 30 runs across six DFJSS scenarios.}
\label{table: test}
\footnotesize
\centering
\begin{tabular}{c|c|c}
\hline
\textbf{Scenario}                     & \textbf{GP}    & \textbf{EvoSpeak}            \\ \hline
\textless{}Tmax, 0.85\textgreater{}   & 858.72(51.49)  & 868.74(48.62)(=)          \\
\textless{}Tmax, 0.95\textgreater{}   & 1050.41(69.10) & 1040.05(57.40)(=)         \\
\textless{}Fmean, 0.85\textgreater{}  & 571.00(12.09)  & 564.81(3.76)($\uparrow$)  \\
\textless{}Fmean, 0.95\textgreater{}  & 624.70(12.26)  & 621.87(3.95)($\uparrow$)  \\
\textless{}WTmean, 0.85\textgreater{} & 450.03(13.44)  & 445.76(10.01)(=)          \\
\textless{}WTmean, 0.95\textgreater{} & 569.96(31.01)  & 559.13(13.86)($\uparrow$) \\ \hline
\end{tabular}
\end{table}

As observed, EvoSpeak significantly outperforms GP in 3 out of 6 scenarios (\textless{}Fmean, 0.85\textgreater{}, \textless{}Fmean, 0.95\textgreater{}, \textless{}WTmean, 0.95\textgreater{}). In the other three scenarios (\textless{}Tmax, 0.85\textgreater{}, \textless{}Tmax, 0.95\textgreater{}, \textless{}WTmean, 0.85\textgreater{}), EvoSpeak demonstrates comparable performance to GP, but without statistical significance. For the four scenarios focusing on mean-based objectives (mean-flowtime and mean-weighted-tardiness), EvoSpeak either significantly outperforms GP, achieving a lower mean test performance with less variability, or at least shows a lower mean test performance with reduced variability, confirming its statistical superiority in these cases. For the two scenarios focusing on max-based objectives (max-tardiness), EvoSpeak performs comparably to GP. Specifically, in \textless{}Tmax, 0.95\textgreater{}, EvoSpeak achieves both a lower mean and standard deviation, indicating slightly better performance.
In \textless{}Tmax, 0.85\textgreater{}, EvoSpeak exhibits a lower standard deviation, suggesting more stable results despite a slightly higher mean. Across all scenarios, EvoSpeak consistently has a lower standard deviation than GP, demonstrating greater stability in performance when using a warm-start population generated by LLM.

The reason EvoSpeak performs better on mean-objectives than on max-objectives is likely because its initial population is derived from effective scheduling heuristics for mean-weighted-tardiness. In DFJSS, mean-flowtime is a more similar task to mean-weighted-tardiness, whereas max-tardiness is fundamentally different, making direct knowledge transfer less effective. These results not only highlight the advantages of EvoSpeak in generating warm-start populations for the same task, but also demonstrate its effectiveness in transferring knowledge to similar tasks, verifying its potential in multi-task optimization \cite{gupta2015multifactorial}. Overall, this suggests that EvoSpeak is effective in learning and generating heuristics that improve scheduling objectives, particularly for tasks that are the same or closely related.

\subsection{Multi-Objective Scenarios}
EvoSpeak’s performance in multi-objective DFJSS scenarios evaluates its ability to generate heuristics aligned with user-defined preference weights while generalizing across unseen preferences. Both standard multi-objective GP and EvoSpeak are trained under three preference configurations: (0.2, 0.8), (0.5, 0.5), and (0.8, 0.2). For instance, GP28 refers to GP trained with preference (0.2, 0.8), while EvoSpeak82 indicates EvoSpeak trained with preference (0.8, 0.2). Each trained model is then tested across all preference settings to evaluate adaptability and generalization.



\subsubsection{Test Performance under Preference (0.2, 0.8)}

\begin{table*}[t]
\caption{The mean and standard deviation test performance of 30 independent runs of GP and EvoSpeak with different preferences across 4 scenarios under preference (0.2, 0.8).}
\label{table:28}
\centering
\footnotesize
\begin{tabular}{c|c>{\columncolor[HTML]{FFFFC7}}c|cc|cc}
\hline
\textbf{Scenario} & \textbf{GP28} & \textbf{EvoSpeak28} & \textbf{GP55} & \textbf{EvoSpeak55} & \textbf{GP82} & \textbf{EvoSpeak82} \\ \hline
\textless{}Fmean-WTmean, 0.85\textgreater{} & 0.617(0.032)($\uparrow$) & 0.608(0.012) & 0.620(0.026)($\uparrow$) & 0.613(0.008)($\uparrow$) & 0.621(0.014)($\uparrow$) & 0.619(0.010)($\uparrow$) \\
\textless{}Tmean-WFmean, 0.85\textgreater{} & 0.746(0.019)(=) & 0.745(0.009) & 0.745(0.011)(=) & 0.743(0.005)($\downarrow$) & 0.746(0.008)(=) & 0.744(0.010)(=) \\
\textless{}Fmean-WTmean, 0.95\textgreater{} & 0.698(0.041)($\uparrow$) & 0.681(0.012) & 0.702(0.050)($\uparrow$) & 0.683(0.011)(=) & 0.697(0.033)($\uparrow$) & 0.694(0.010)($\uparrow$) \\
\textless{}Tmean-WFmean, 0.95\textgreater{} & 0.783(0.013)(=) & 0.781(0.009) & 0.794(0.032)($\uparrow$) & 0.782(0.007)(=) & 0.789(0.020)($\uparrow$) & 0.787(0.006)($\uparrow$) \\ \hline
    Win|draw|lose            &     2|2|0    &       -      &     3|1|0    &   1|2|1      &   3|1|0      &  3|1|0             \\  \hline
Average rank         &       4.0      &       \textbf{1.75}        &       5.00     &        1.75       &      5.25      &       3.25        \\ \hline
\end{tabular}
\end{table*}

Table \ref{table:28} presents the mean and standard deviation of test performance from 30 independent runs of GP and EvoSpeak across four scenarios under preference (0.2, 0.8). The table evaluates test performance under this preference while considering different training preferences. In the comparisons, EvoSpeak28 is tested against each method, with the symbols ($\uparrow$), ($\downarrow$), and (=) indicating whether it performs significantly better, worse, or shows no statistical difference, based on the Wilcoxon test at a 0.05 significance level. Additionally, the Friedman test \cite{zimmerman1993relative} is conducted to rank these methods. The results show that EvoSpeak28 generally outperforms or matches other methods across all scenarios, except against EvoSpeak55 in \textless{}Tmean-WFmean, 0.85\textgreater{}. This confirms that LLM-generated populations enhance the search process and exhibit strong generalization to new testing conditions. Notably, EvoSpeak28 significantly outperforms GP28 in multiple scenarios, and the Friedman test ranks it highest under preference (0.2, 0.8), reinforcing the effectiveness of LLM-initialized individuals in guiding evolution. Even when trained under different preferences, EvoSpeak82 and EvoSpeak55 adapt well to testing under preference (0.2, 0.8), often outperforming or matching methods trained under the same preference. This suggests that LLM-generated individuals capture structural patterns that remain effective across varying test conditions. The average rank of 1.75 for EvoSpeak28 further highlights its superior performance and supports the role of LLM-generated individuals in improving evolutionary learning and aligning with user preference.

\subsubsection{Test Performance under Preference (0.5, 0.5)}

\begin{table*}[t]
\caption{The mean and standard deviation test performance of 30 independent runs of GP and EvoSpeak with different preferences across 4 scenarios under preference (0.5, 0.5).}
\label{table:55}
\centering
\footnotesize
\begin{tabular}{c|cc|c>{\columncolor[HTML]{FFFFC7}}c|cc}
\hline
\textbf{Scenario} & \textbf{GP28} & \textbf{EvoSpeak28} & \textbf{GP55} & \textbf{EvoSpeak55} & \textbf{GP82} & \textbf{EvoSpeak82} \\ \hline
\textless{}Fmean-WTmean, 0.85\textgreater{} & 0.681(0.026)(=) & 0.675(0.011)(=) & 0.682(0.021)(=) & 0.676(0.007) & 0.681(0.012)($\uparrow$) & 0.680(0.008)($\uparrow$) \\
\textless{}Tmean-WFmean, 0.85\textgreater{} & 0.680(0.023)(=) & 0.680(0.011)($\uparrow$) & 0.677(0.013)($\uparrow$) & 0.673(0.007) & 0.677(0.010)($\uparrow$) & 0.675(0.012)(=) \\
\textless{}Fmean-WTmean, 0.95\textgreater{} & 0.747(0.035)($\uparrow$) & 0.734(0.010)(=) & 0.748(0.042)($\uparrow$) & 0.731(0.009) & 0.741(0.027)($\uparrow$) & 0.737(0.009)($\uparrow$) \\
\textless{}Tmean-WFmean, 0.95\textgreater{} & 0.735(0.016)($\uparrow$) & 0.733(0.011)($\uparrow$) & 0.744(0.039)($\uparrow$) & 0.728(0.007) & 0.736(0.025)($\uparrow$) & 0.732(0.007)($\uparrow$) \\ \hline
    Win|draw|lose    &     2|2|0    &    2|2|0     &     3|1|0    &   -     &   4|0|0      &  3|1|0             \\  \hline
Average rank         &       4.5      &       3.00        &       5.25     &        \textbf{1.25}       &      4.5      &       2.5        \\ \hline
\end{tabular}
\end{table*}

Similarly, Table \ref{table:55} presents the test performance under preference (0.5, 0.5), comparing EvoSpeak55 with other methods using the same Wilcoxon test significance indicators. The results show that EvoSpeak55 consistently performs better or on par with other methods, confirming the effectiveness of LLM-guided initialization under this preference. Despite differences in training preferences, EvoSpeak82 and EvoSpeak55 demonstrate strong adaptability, often outperforming or matching methods trained under the same preference. This further supports the hypothesis that LLM-generated individuals capture structural patterns that generalize well. The average ranking of 1.25 for EvoSpeak55 under preference (0.5, 0.5) reinforces its advantage in evolutionary learning and alignment with user preference.

\subsubsection{Test Performance under Preference (0.8, 0.2)}
\begin{table*}[t]
\caption{The mean and standard deviation test performance of 30 independent runs of GP and EvoSpeak with different preferences across 4 scenarios under preference (0.8, 0.2).}
\label{table:82}
\centering
\footnotesize
\begin{tabular}{c|cc|c>{\columncolor[HTML]{EFEFEF}}c|c>{\columncolor[HTML]{FFFFC7}}c}
\hline
\textbf{Scenario} & \textbf{GP28} & \textbf{EvoSpeak28} & \textbf{GP55} & \textbf{EvoSpeak55} & \textbf{GP82} & \textbf{EvoSpeak82} \\ \hline
\textless{}Fmean-WTmean, 0.85\textgreater{} & 0.745(0.020)(=) & 0.742(0.009)(=) & 0.745(0.016)(=) & 0.738(0.005)($\downarrow$) & 0.742(0.009)(=) & 0.741(0.007) \\
\textless{}Tmean-WFmean, 0.85\textgreater{} & 0.615(0.027)(=) & 0.615(0.013)($\uparrow$) & 0.608(0.016)(=) & 0.604(0.008)(=) & 0.608(0.012)(=) & 0.606(0.014) \\
\textless{}Fmean-WTmean, 0.95\textgreater{} & 0.797(0.029)($\uparrow$) & 0.787(0.010)($\uparrow$) & 0.794(0.035)($\uparrow$) & 0.779(0.007)(=) & 0.786(0.021)(=) & 0.781(0.007) \\
\textless{}Tmean-WFmean, 0.95\textgreater{} & 0.686(0.020)(=) & 0.685(0.015)($\uparrow$) & 0.693(0.046)($\uparrow$) & 0.673(0.008)($\downarrow$) & 0.682(0.029)(=) & 0.678(0.009) \\ \hline
    Win|draw|lose    &     1|3|0    &    3|1|0     &     2|2|0    &   0|2|2     &   0|4|0      &  -             \\  \hline
Average rank         &       5.5      &       4.25        &       5.0     &        \textbf{1.0}       &      3.25      &       \textbf{2.0}        \\ \hline
\end{tabular}
\end{table*}

\begin{figure}[t]
\centerline{\includegraphics[width=0.5\textwidth]{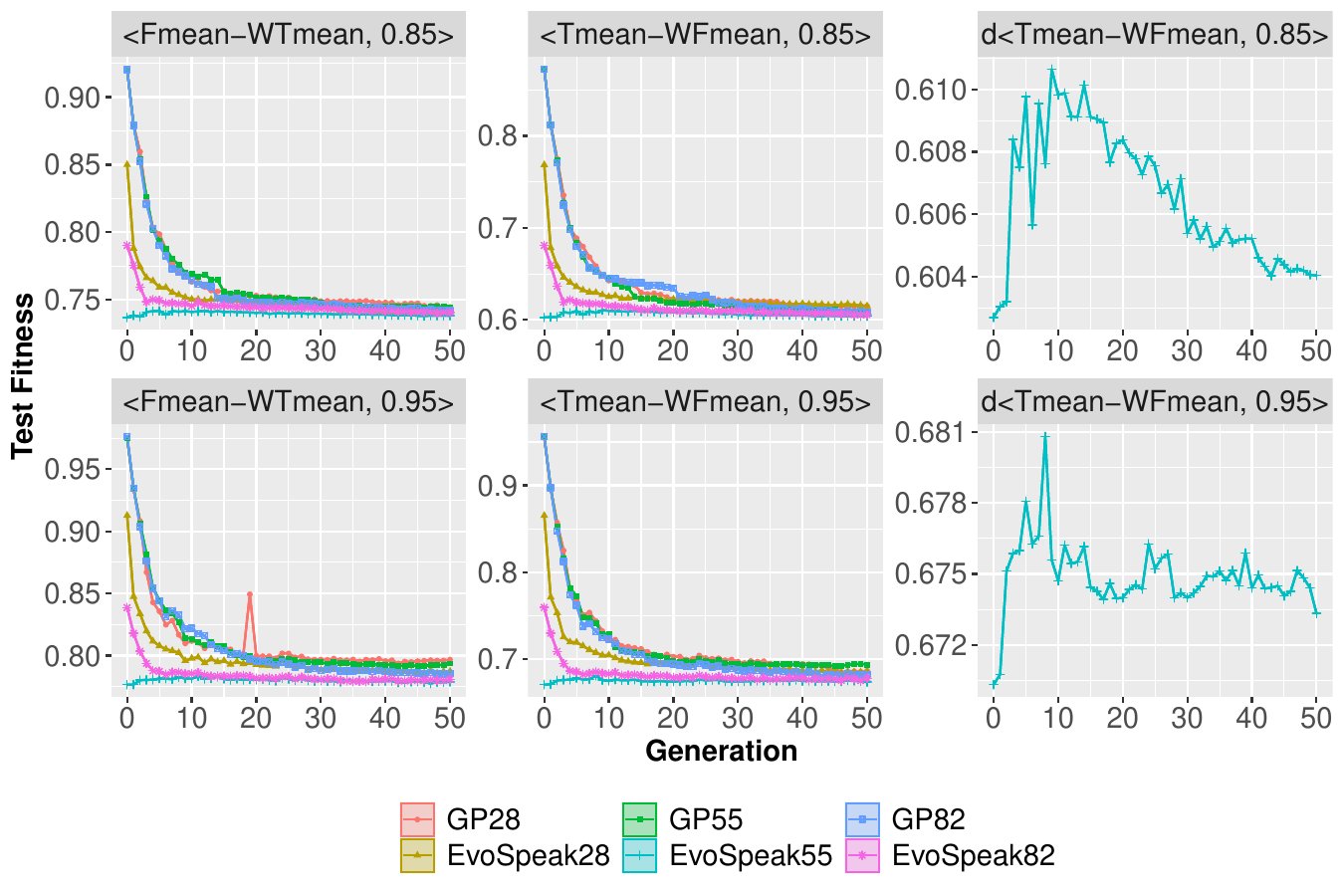}}
\caption{The convergence curves of test performance of 30 independent runs of GP and EvoSpeak with different preferences across 4 scenarios under preference (0.8, 0.2).}
\label{fig: convergence}
\end{figure}

Table \ref{table:82} demonstrates test performance under preference (0.8, 0.2), confirming EvoSpeak's consistent superiority over GP. While Wilcoxon tests reveal EvoSpeak82 often winning or drawing against GP across preferences, it trails EvoSpeak55 in two scenarios and performs similarly in the others, resulting in a ranking deviation (EvoSpeak55: 1.0, EvoSpeak82: 2.0). Fig. \ref{fig: convergence} provides further insight through convergence curve visualization. Notably, EvoSpeak82 exhibits a superior initial population compared to all GP and most EvoSpeak variants, except EvoSpeak55. This suggests that the LLM, in some instances, may generate exceptionally robust individuals that perform well across multiple preference sets, leading to the observed anomaly. 

However, when EvoSpeak55 is excluded from consideration, EvoSpeak82 consistently delivers the best overall performance, validating the efficacy of LLM-guided initialization in aligning with specified preference settings. The slight performance degradation observed in EvoSpeak55's later generations, as depicted in the right subFig.s of Fig. \ref{fig: convergence}, indicates that while the LLM can produce super-strong initial individuals, subsequent GP evolution via subtree modifications can introduce minor performance fluctuations. Nevertheless, the convergence trends suggest that with extended generations, EvoSpeak55 has the potential to surpass its initial high performance, potentially demonstrating the generalizability of these individuals.

Overall, the experimental results strongly validate the effectiveness of LLM-generated initial individuals in enhancing GP-based heuristic learning for DFJSS scheduling. By providing high-quality initial heuristics, LLM reduces early-generation randomness, leading to a more efficient and robust evolutionary process. Moreover, it successfully incorporates user-defined preferences, enabling greater flexibility in heuristic learning. Despite variations in training preferences, EvoSpeak demonstrates strong generalization to unseen test conditions, underscoring its adaptability and robustness. These findings highlight LLM-assisted GP as a promising approach for intelligent heuristic learning in complex scheduling problems, paving the way for further advancements in AI-driven manufacturing optimization.

\section{Further Analysis}
\label{further}


\subsection{Initial Population Terminal Use Frequency}

\begin{figure}[t]
\centerline{\includegraphics[width=0.44\textwidth]{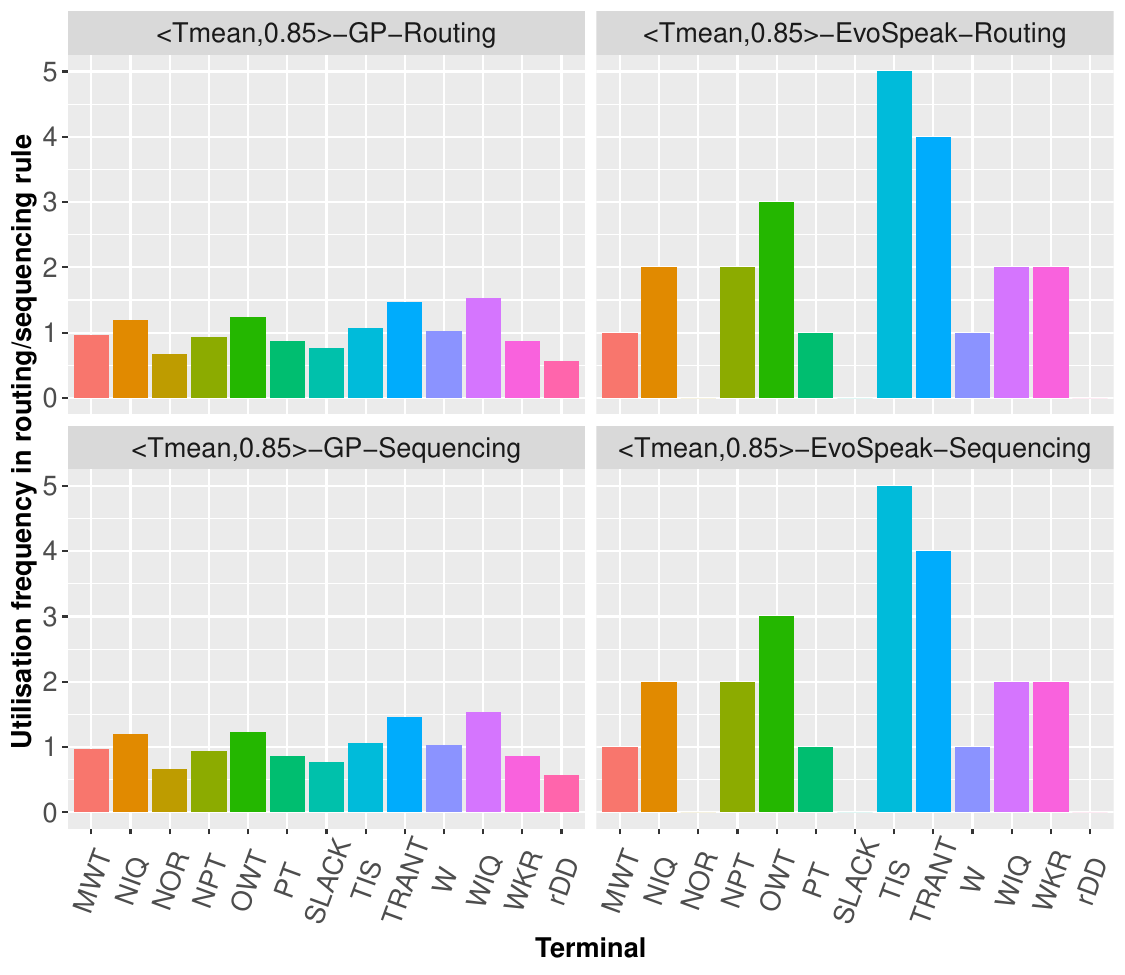}}
\caption{Average terminal usage frequency of the best routing and sequencing rules in the initial population, generated by baseline GP and EvoSpeak, under the scenario \textless Tmean, 0.85\textgreater{} across 30 runs.}
\label{fig: terminalFrequency}
\end{figure}

Fig.~\ref{fig: terminalFrequency} illustrates the terminal usage frequency of the best routing and sequencing rules from the initial population, generated by both the baseline GP and EvoSpeak methods, under the scenario \textless Tmean, 0.85\textgreater{} across 30 independent runs. For EvoSpeak, some high-quality heuristics for mean-flowtime objectives are provided and EvoSpeak will gain knowledge from these heuristics, which may influence the terminal preferences observed.

As shown, the terminal usage in GP-generated rules appears relatively uniform across different terminals for both routing and sequencing rules. This uniformity is expected, as GP begins with a randomly initialized population without incorporating prior domain knowledge. Minor variations are present due to the selection of the best-performing individual in the initial population, rather than analyzing the entire population. In contrast, EvoSpeak exhibits a clear bias toward certain terminals, which is expected since EvoSpeak leverages prior knowledge extracted from existing high-quality heuristics. Consequently, EvoSpeak tends to generate rules that selectively emphasize a subset of influential terminals. Notably, EvoSpeak avoids using certain terminals such as SLACK, rDD, and NOR. These terminals are typically more relevant to objectives involving tardiness rather than mean flowtime, and their exclusion indicates that EvoSpeak has internalized task-specific relevance from prior experience. Moreover, EvoSpeak demonstrates a strong preference for terminals such as TIS (time in system) and TRANT (transportation time), which are likely more informative in minimizing mean flowtime in this specific scenario. Interestingly, this bias is consistent across both routing and sequencing rules, suggesting that EvoSpeak not only encodes task-relevant features but also reuses them consistently across rule types. Another observation is that EvoSpeak tends to produce rules with higher overall terminal usage counts than GP. This may be attributed to the fact that EvoSpeak-generated heuristics are often more complex and longer in structure, as they are synthesized from previously learned rule patterns that may inherently be more elaborate.

Overall, the comparison highlights the different behaviors of the two methods: GP explores terminals more uniformly due to random initialization, while EvoSpeak introduces informed biases by exploiting prior knowledge, thereby producing heuristics that more closely align with task-specific characteristics from the outset. This difference is expected to influence the subsequent evolutionary search dynamics and the diversity of solutions discovered.

\subsection{Population Diversity}

\begin{figure}[t]
\centerline{\includegraphics[width=0.44\textwidth]{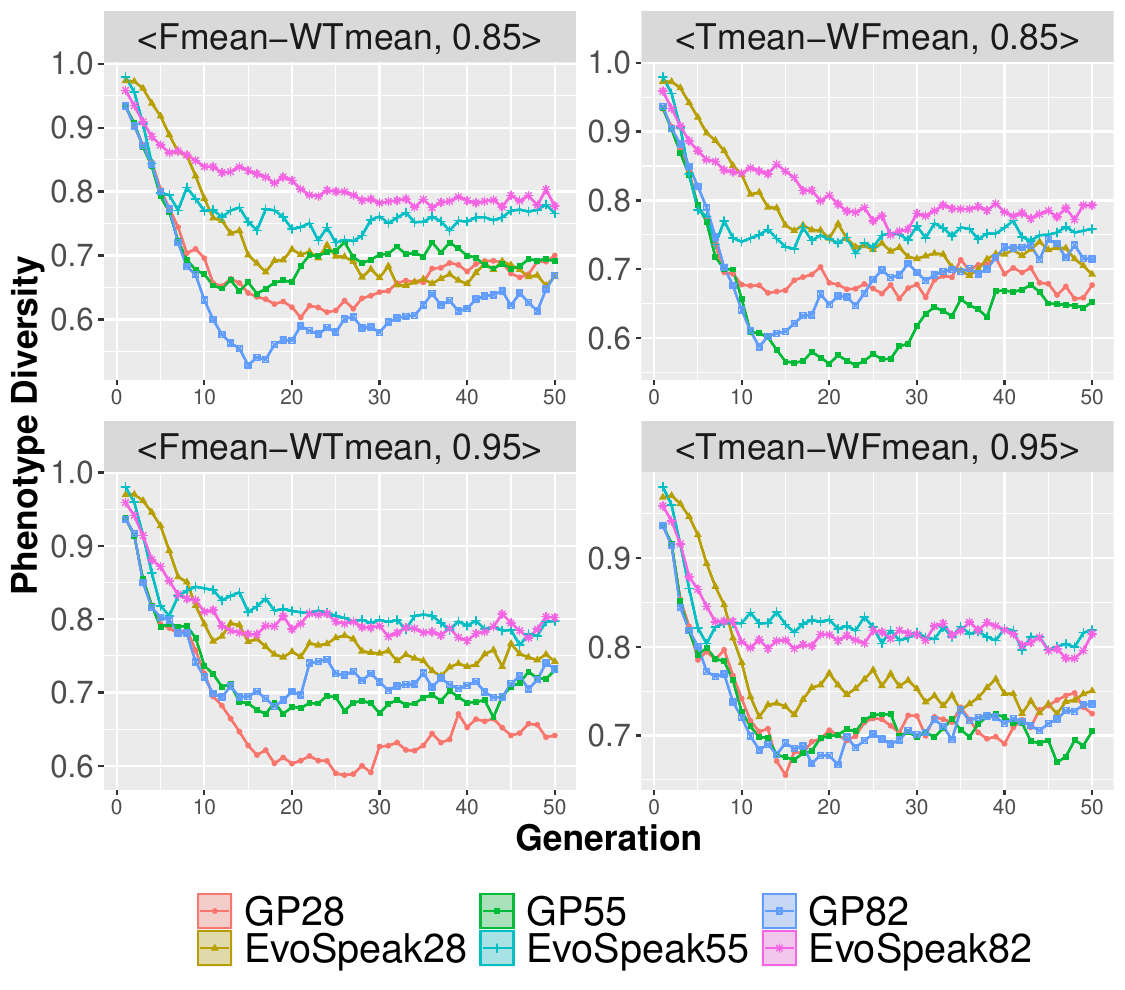}}
\caption{Average phenotypic diversity of the population over 30 runs during the evolutionary process of baseline GP and EvoSpeak under different user preferences across four multi-objective scenarios.}
\label{fig: diversity}
\end{figure}

Fig.~\ref{fig: diversity} shows the average phenotypic diversity of the population over 30 runs during the evolutionary process of baseline GP and EvoSpeak under different user preferences across four multi-objective scenarios. Here, phenotypic diversity is defined as the percentage of individuals with unique fitness values within the population. Compared with genotypic diversity, which focuses on structural variation, phenotypic diversity better reflects the behavioral richness of the evolved heuristics and their ability to explore different regions of the objective space.

At the beginning of the evolutionary process, all methods exhibit high diversity with no substantial differences. However, as evolution progresses, a clear divergence emerges. Across all scenarios, baseline GP rapidly loses diversity, with populations converging prematurely toward limited regions of the search space. This collapse in diversity restricts GP’s ability to sustain exploration, increasing the risk of stagnation at suboptimal solutions. In contrast, EvoSpeak consistently maintains significantly higher phenotypic diversity throughout the evolutionary process. This advantage can be attributed to two key factors:
\begin{enumerate}
    \item Knowledge-guided initialization: EvoSpeak synthesizes heuristics by leveraging prior knowledge from high-performing heuristics. These heuristics are typically more complex and structurally richer, which results in broader initial behavioral variability compared to the random trees generated by GP.
    \item User-preference-driven prompting: In addition to reusing prior knowledge, EvoSpeak integrates user preferences into the heuristic generation process by instructing the LLM to generate a more diverse set of heuristics for the initial population. This dual guidance not only aligns the search with user-specified objectives but also enriches the pool of building blocks, thereby mitigating early homogenization of the population.
\end{enumerate}

As a result, even in later generations, EvoSpeak maintains higher levels of diversity across all user-preference settings and scenarios. Maintaining higher diversity offers two critical advantages. First, it improves the robustness of search by enabling populations to adapt to different trade-offs in multi-objective optimization. Second, it reduces the risk of premature convergence, thereby improving the chances of discovering high-quality heuristics over time. These results demonstrate that EvoSpeak not only improves convergence speed and heuristic quality but also fundamentally enhances the evolutionary dynamics by sustaining a healthier exploration–exploitation balance compared to baseline GP.

\subsection{From Raw GP Tree to Natural EvoSpeak Rule}


\begin{figure}[t]
\centerline{\includegraphics[width=0.49\textwidth]{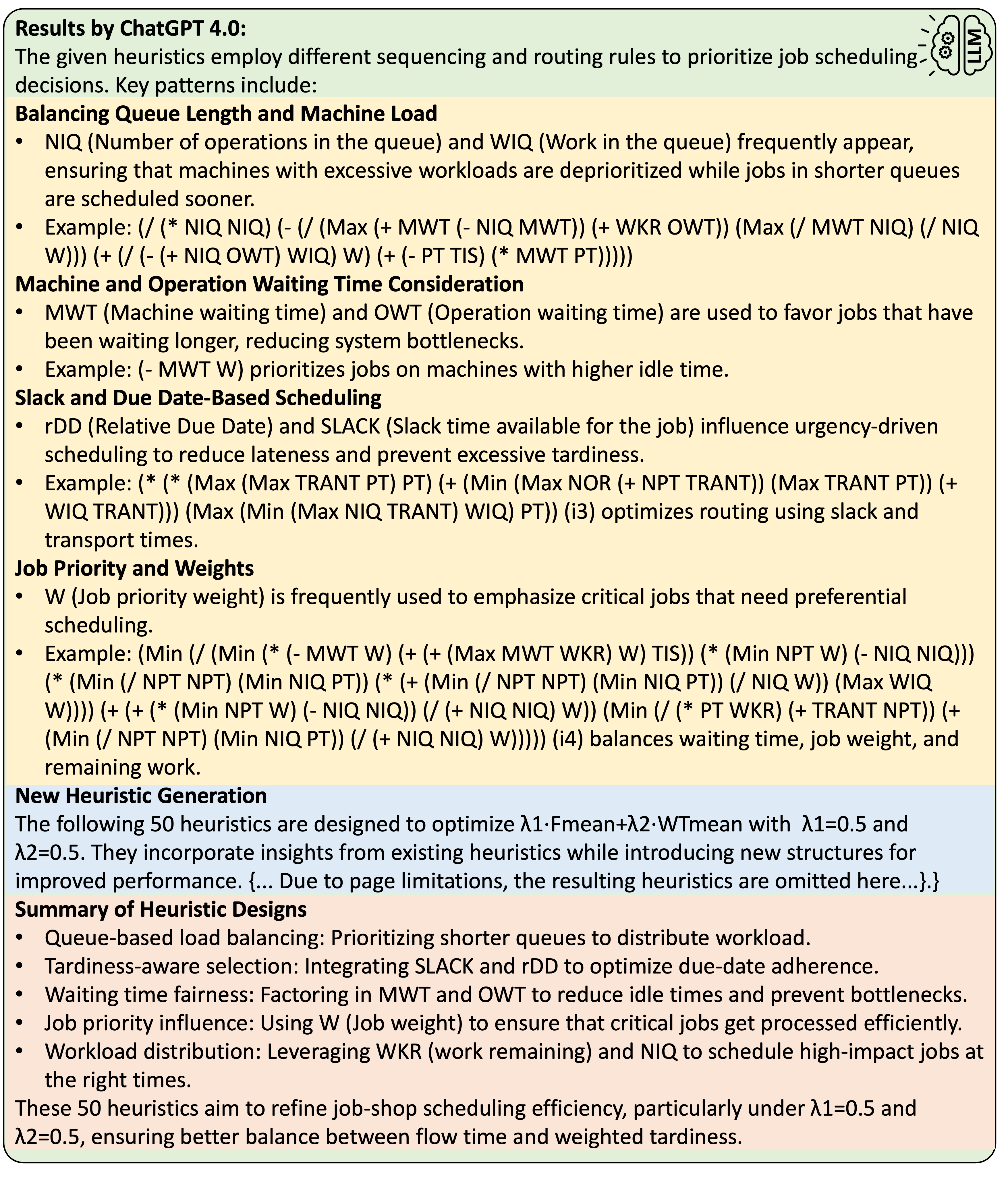}}
\caption{An example result by ChatGPT 4.0 for heuristics analysis and population initialization considering user preferences for a multi-objective DFJSS problem.}
\label{fig: LLMresults}
\end{figure}

Fig.~\ref{fig: LLMresults} illustrates an example output from EvoSpeak, highlighting its capability to analyze existing scheduling heuristics and generate a high-quality initial population for multi-objective DFJSS problems. Given a set of pre-evolved heuristics and explicit user preference information, EvoSpeak processes the input to identify key structural patterns, operational dependencies, and performance trade-offs between competing objectives. This analysis enables the extraction of meaningful knowledge from complex heuristics, which can then be transformed into new candidate heuristics that respect both the underlying patterns and the specified user preferences.

By leveraging the insights obtained from existing heuristics, EvoSpeak generates a population that is already aligned with the optimization objectives, providing a warm start for subsequent GP evolution. This warm-start population not only accelerates convergence but also improves the stability and quality of evolved heuristics, reducing the randomness typically observed in early generations. Furthermore, EvoSpeak produces human-readable explanations of the generated heuristics, enhancing interpretability and enabling users—especially those without deep domain expertise—to understand the rationale behind scheduling decisions.

\section{Conclusions}
\label{conclusion}
This paper proposes EvoSpeak, a novel framework that integrates LLMs with GP to enhance the evolution, interpretability, and transferability of heuristics for complex scheduling and optimization problems. By leveraging the symbolic reasoning and knowledge extraction capabilities of LLMs, EvoSpeak can analyze existing high-performing heuristics, uncover their underlying structural patterns, and generate a warm-start population that accelerates GP evolution. This synergy not only improves convergence efficiency but also enables effective transfer learning across related tasks, allowing the system to adapt knowledge from one scenario to another with minimal additional training. Extensive experimental evaluation on both single- and multi-objective DFJSS problems demonstrates that EvoSpeak consistently outperforms traditional GP approaches. The LLM-initialized populations yield faster convergence, higher-quality final heuristics, and reduced variability across runs. Moreover, EvoSpeak effectively incorporates user-defined preferences into heuristic generation, producing heuristics aligned with multi-objective trade-offs. Importantly, the framework provides human-readable interpretations of evolved heuristics, enhancing transparency, facilitating domain understanding, and building trust in automated decision-making processes. Despite these strengths, EvoSpeak’s performance remains dependent on the quality of LLM outputs, which may introduce biases or inaccuracies. Nonetheless, with the rapid advancement of LLMs, this reliance is expected to further strengthen EvoSpeak’s capabilities rather than hinder them.

The proposed framework also opens several avenues for future research. First, EvoSpeak’s LLM–GP integration can be extended to other combinatorial and continuous optimization domains, such as vehicle routing, production planning, or inventory management, to evaluate the generality of knowledge transfer. Second, interactive LLM-powered interfaces could enable real-time user guidance, allowing practitioners to steer heuristic generation and interpretation dynamically. Finally, applying EvoSpeak to large-scale, real-world optimization problems—such as supply chain networks, logistics planning, or financial portfolio optimization—would further validate its practical impact and demonstrate its potential for industry adoption.

In summary, EvoSpeak represents a significant step toward evolving heuristics that are not only efficient and adaptable but also interpretable and preference-aware. By bridging human-understandable reasoning with automated evolutionary search, it provides a powerful and transparent tool for tackling complex optimization challenges across both research and industrial applications.


%



\section*{Acknowledgment}
The authors gratefully acknowledge the assistance of AI-based language tools, including ChatGPT and Gemini, which were used to refine the writing and enhance the readability of this paper. Typical prompts involved requests such as ``please refine and revise the following content for a paper.''


\ifCLASSOPTIONcaptionsoff
  \newpage
\fi



%



\bibliographystyle{IEEEtran}
\bibliography{reference}
\end{document}